\documentclass[wcp]{jmlr}
\usepackage{times}
\usepackage{latexsym}
\usepackage{hyperref}
\usepackage{url}

\usepackage[utf8]{inputenc} 
\usepackage[T1]{fontenc}    
\usepackage{booktabs}       
\usepackage{amsfonts}       
\usepackage{nicefrac}       
\usepackage{microtype}      
\usepackage{amsmath,amssymb}
\usepackage{graphicx}
\usepackage{natbib}
\usepackage[export]{adjustbox}
\usepackage{subcaption}
\usepackage{placeins}
\newcommand*{\doi}[1]{\href{http://dx.doi.org/\detokenize{#1}}{doi: \detokenize{#1}}}

\graphicspath{{./},{./figures/}}

\jmlrvolume{80}
\jmlryear{2017}
\jmlrworkshop{ACML 2017}

\title{Nested LSTMs}

\author{\Name{Joel Ruben Antony Moniz} \Email{jrmoniz@andrew.cmu.edu} \\
\addr Carnegie Mellon University\thanks{Work begun while the author was at MILA and \'Ecole Polytechnique de Montr\'eal}
\AND
\Name{David Krueger} \Email{david.krueger@umontreal.ca}\\
\addr MILA, Universit\'e de Montr\'eal
 }

\begin{document}
\maketitle


\begin{abstract}
We propose \emph{Nested LSTMs} (NLSTM), a novel RNN architecture with multiple levels of memory.
Nested LSTMs add depth to LSTMs via nesting as opposed to stacking.
The value of a memory cell in an NLSTM is computed by an LSTM cell, which has its own {\it inner} memory cell.
Specifically, instead of computing the value of the (outer) memory cell as $c^{outer}_t = f_t \odot c_{t-1} + i_t \odot g_t$, NLSTM memory cells use the concatenation $(f_t \odot c_{t-1}, i_t \odot g_t)$ as input to an inner LSTM (or NLSTM) memory cell, and set $c^{outer}_t$ = $h^{inner}_t$.
Nested LSTMs outperform both stacked and single-layer LSTMs with similar numbers of parameters in our experiments on various character-level language modeling tasks, and the inner memories of an LSTM learn longer term dependencies compared with the higher-level units of a stacked LSTM.
\end{abstract}


\section{Introduction}

Learning long-term dependencies is a key challenge for current machine learning approaches to artificial intelligence.
The ability of human beings to reconcile these long-term dependencies with the immediate context, i.e., to adapt and use knowledge that has been previously gained so as to be relevant to the current frame-of-reference, is indispensable. 
An important example of this ability, if on a much smaller scale, is the ability to predict characters and words in a sentence or document based on one's past experience (for example, in the form of commonly encountered constructions and phrases), the general subject dealt with in the document, and the precise wording of the specific sentence in question. 
Recurrent neural network based architectures have made significant progress towards having a machine mimic this ability.

Recurrent neural networks (RNNs) condition their present representation of the state of the world on their entire history of inputs (or ``observations'' in reinforcement learning parlance), and so are a natural fit for learning temporally abstracted features.
In theory, a simple RNN can represent arbitrary functions and thus have the capacity to solve tasks involving dependencies at arbitrary time-scales.
In practice, more complex architectures have proven essential for solving many tasks. 
One reason for this is the vanishing gradient problem \citep{hochreiter1991,vanishing_gradient}, which makes it difficult for simple RNNs to learn long-term dependencies.
Successful RNN architectures, such as LSTMs \citep{LSTM} typically incorporate memory mechanisms which ameliorate the problem of vanishing gradient. 

A more fundamental issue is that learning to detect long-term dependencies involves a fundamentally difficult credit assignment problem: in the absence of prior information, any past event may plausibly be responsible for current events.
Architectural features such as memory mechanisms encode implicit priors which may help with the credit assignment problem.
Memory mechanisms allow a model to remember past information over arbitrarily long time-scales, so that credit can be assigned to events in the distant past.
We seek to encode an additional implicit prior of {\it temporal hierarchy} by the creation of a novel memory structure. 
In particular, we suggest selective memory access via nesting as an approach to constructing temporal hierarchies in memory.

While some prior work on hierarchical memory exists, LSTM (and variants) are still the most 
popular deep learning model for sequential tasks, such as in character-level language modeling. 
In particular, the default Stacked LSTM architecture uses a sequence of LSTMs stacked on top of each other to process the data, the input to a layer being the output of the previous layer. 
In this work, we propose and explore a novel {\it Nested} LSTM architecture (NLSTM), which we envision as a potential drop-in replacement for a stacked LSTM. 

In NLSTMs, the LSTM memory cells have access to an {\it inner} memory, which they selectively read and write to using the standard LSTM gates.
This key feature allows the model to implement a more effective temporal hierarchy than a conventional Stacked LSTM.
In NLSTM, the (outer) memory cell are free to selectively read and write relevant long-term information to their inner cell.
In contrast, in stacked LSTMs, the upper-level activations (analogous to the inner memories) are directly accessed to produce an output, and therefore must contain all the short-term information which is relevant to the current prediction.
In other words, the primary difference between stacked LSTMs and Nested LSTMs is the idea of selective access to inner memories which the NLSTM implements.
This frees the inner memories to remember and process events on longer time scales, even when these events are not relevant to the immediate present.

Our visualizations demonstrate that the inner memories of NLSTMs do in fact operate on longer time-scales than higher-level memories in a stacked LSTM.
Our experiments also show that NLSTMs outperform Stacked LSTMs in a wide range of tasks.

\section{Related Work}

The problem of learning effective temporal hierarchies for dealing with long-term dependencies is well studied in the context of both RNNs and reinforcement learning.
A comprehensive review of this topic is beyond our scope, we review some recent works and focus on the distinctive aspects of our approach.

Doing credit assignment over long time-scales is a central problem of reinforcement learning.
The options framework in RL \citep{sutton1999between} enables long-term planning over sequences of temporally abstracted actions called options.
Selecting an option amounts to temporarily enacting a subpolicy which then selects the primitive actions at each time-step (or its own options).
Although learning options has received some attention \citep{stolle2002learning, brunskill2014pac}, including some recent gradient-based approaches \citep{options_heads, option_critic}, most successful applications so far have used hand-crafted options.

\subsection{Deep learning approaches to temporal abstraction}

Currently, RNNs are frequently \emph{stacked} creating a multi-layer feedforward network at each time-step.
\citet{recurrent_depth0} argue that stacking may results in more abstract, long-term features; \citet{recurrent_depth} argue that this may not be the case.
Unlike stacking, nesting also increases recurrent depth, which can improve performance \citet{recurrent_depth}.
\citet{deepRNN} add multi-layer input, output, or recurrent connections as an alternative to stacking; their deep recurrent connections increase recurrent depth, but are not commonly used.
Multi-layer input connections have been used, however, for state-of-the-art speech-recognition \citep{deep_speech, deep_speech2} systems; these systems also incorporate stacked RNNs.

Our model is based on the popular Long Short-term Memory (LSTM) \citep{LSTM} architecture.
The hidden states of LSTMs include internal memory cells, which use identity connections to store long-term memories.
The LSTM forget/remember\footnote{
We support the efforts of \citet{fast_weights} to reverse this counter-intuitive naming convention.
} gate \citep{forget_gate} allows memories to be forgotten with an (adaptive) multiplicative decay on these identity connections.

A wide variety of network architectures based on or inspired by LSTMs have been proposed \citep{multidimensionalLSTM, GRU1, GRU2, gridLSTM, ALSTM, memLSTM}.
Perhaps the most popular and well know is the Gated Recurrent Unit (GRU) \citep{GRU1, GRU2}.
GRUs function similarly to LSTMs, but they do not feature any internal memory; the entire hidden state is exposed to external computational units.
This moves in the opposite direction of our work, which is focused on creating \emph{more} internal memory.
Some recent works also apportion more of the total hidden-state into internal memories \citep{memLSTM, arrayLSTM}, but not in a way which involves nesting.
\citet{searchspace,rnn_exploration} evaluate architectural variants of LSTMs and GRUs; \citet{searchspace} remove components of standard LSTMs, whereas \citet{rnn_exploration} use an evolutionary search procedure to search a wider space of possible models.

The LSTM remember gates allow the model to dynamically decay memories of different units at different rates, but  do not explicitly encourage different units to model different levels of temporal dependency. 
Some other works attempt to encode the temporal hierarchy in the prior of a recurrent model.
Temporal hierarchies among units can explicitly coded by hand, as in Clockwork RNNs \citep{koutnik2014clockwork} and hierarchical RNNs \citep{hierarchical_rnn}.
This approach seems brittle; it would be preferable for the model to learn to operate at the appropriate time-scales.
\citet{gfrnn} present a fully differentiable approach to this problem, based on adding additional gating mechanisms. 
A downside of this work is that the model size grows quadratically in the number of layers in the hierarchy. 
More recently, \citet{hmrnn} use the straight-through estimator \citep{straight-through1,straight-through2} to train a model which makes ``crisp'' binary decisions about when to update different recurrent units.

Recent work in Deep Learning considers augmenting RNN architectures with novel memory mechanisms inspired by computer memory architectures \citep{NTM, stack_machine_fb, stack_machine_dm} and neural short-term memory mechanisms \citep{fast_weights}.
Storing and accessing memories provide paths for gradient flow, but, just as in RNNs, using backpropagation through time becomes prohibitively computationally expensive when sequences become long.
The standard solution to this problem is truncate gradient flow after some number of time-steps.  
\citet{RLNTM} attempt to use reinforcement learning (specifically, the REINFORCE algorithm \citep{REINFORCE}) to solve this problem in the context of training Neural Turing Machines (NTMs).

\section{Nested LSTMs}
The output gate in LSTMs encodes the intuition that memories which are not relevant at the present time-step may still be worth remembering.
Nested LSTMs use this intuition to create a temporal hierarchy of memories.
Access to the inner memories is gated in exactly the same way, so that longer-term information which is only situationally relevant can be accessed selectively.

\begin{figure}[h]
    \centering
        \includegraphics[width=0.48\textwidth]{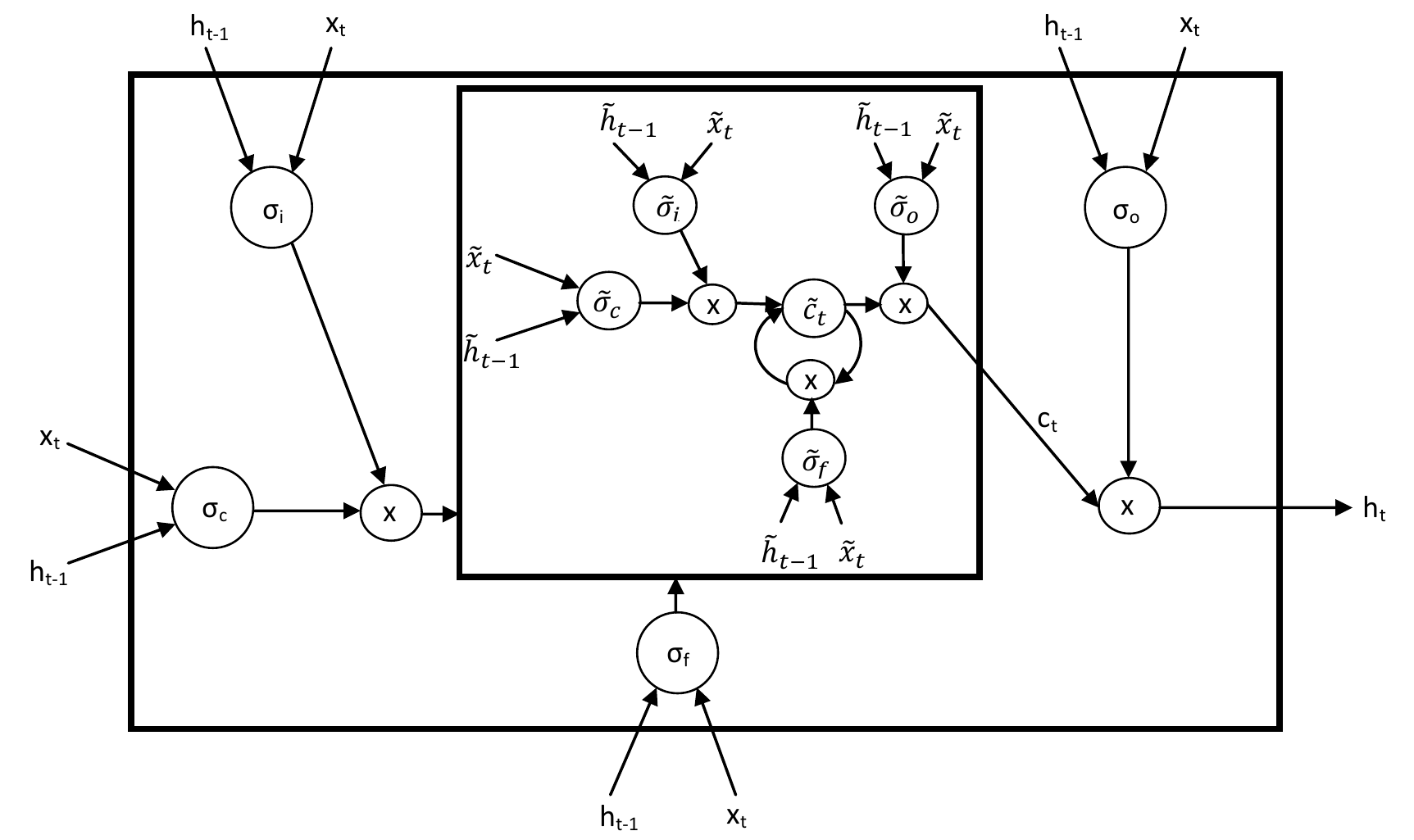}
    \caption{The Nested LSTM architecture}
    \label{fig:nested}
\end{figure}

\begin{figure*}[t]
\centering
\subcaptionbox{LSTM\label{fig:sfig1}}[.3\linewidth]{\includegraphics[clip, trim=2cm 3cm 1cm 3.5cm, width=0.30\textwidth]{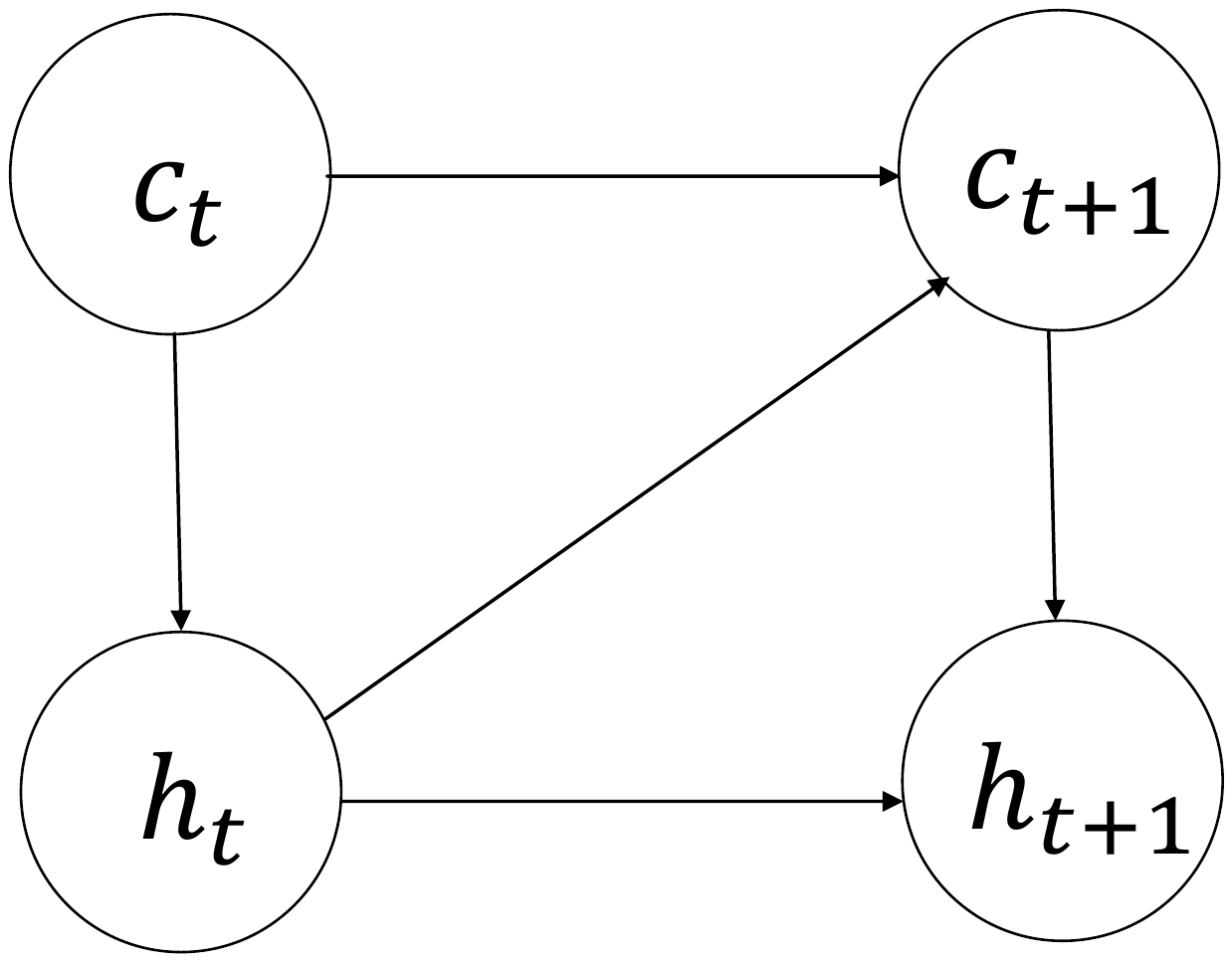}}
\subcaptionbox{Stacked LSTM\label{fig:sfig2}}
  [.3\linewidth]{\includegraphics[clip, trim=2cm 2cm 1cm 2cm, width=0.30\textwidth]{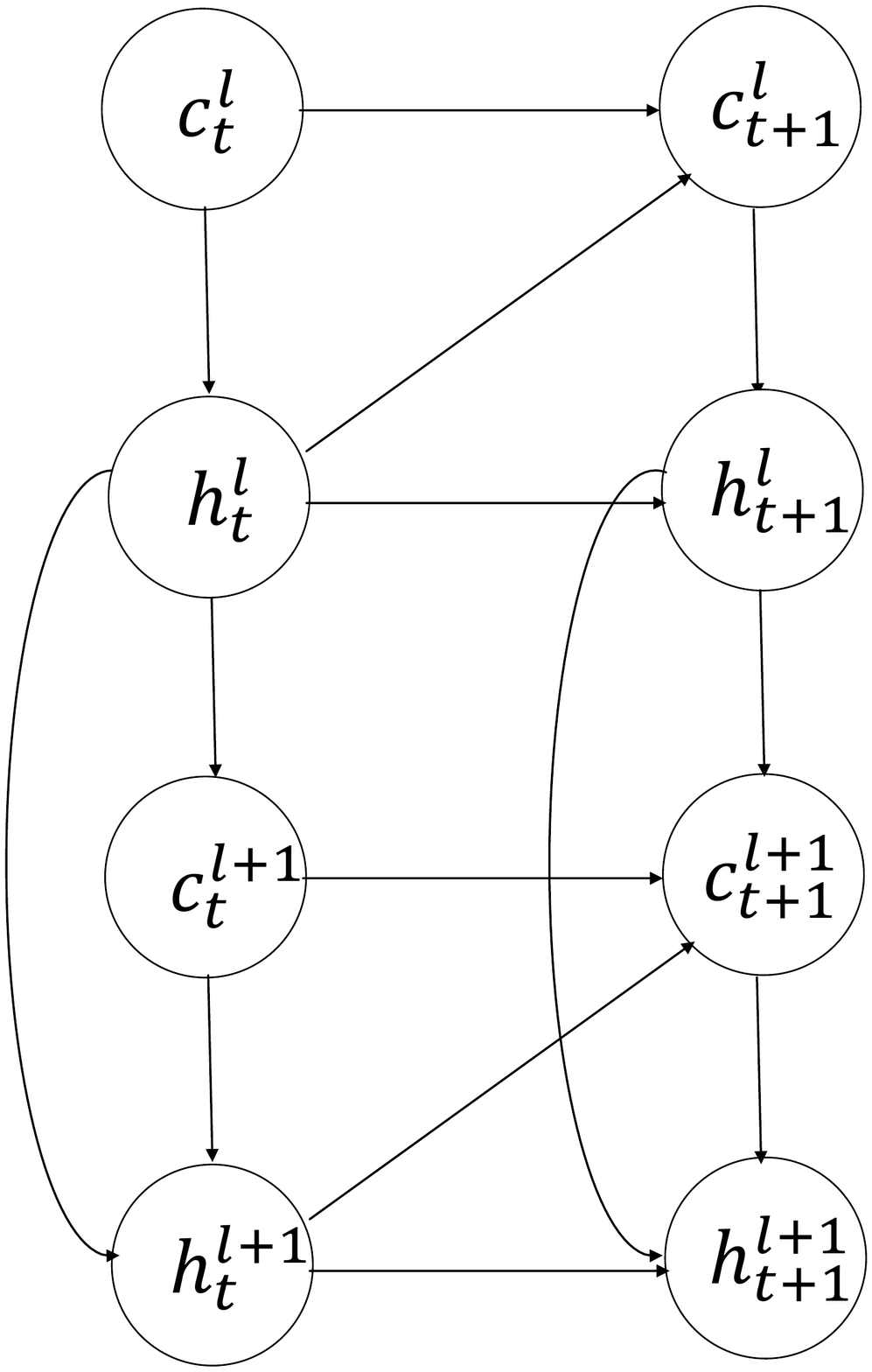}}
\subcaptionbox{Nested LSTM\label{fig:sfig3}}
  [.3\linewidth]{\includegraphics[clip, trim=2cm 2cm 1cm 3cm, width=0.30\textwidth]{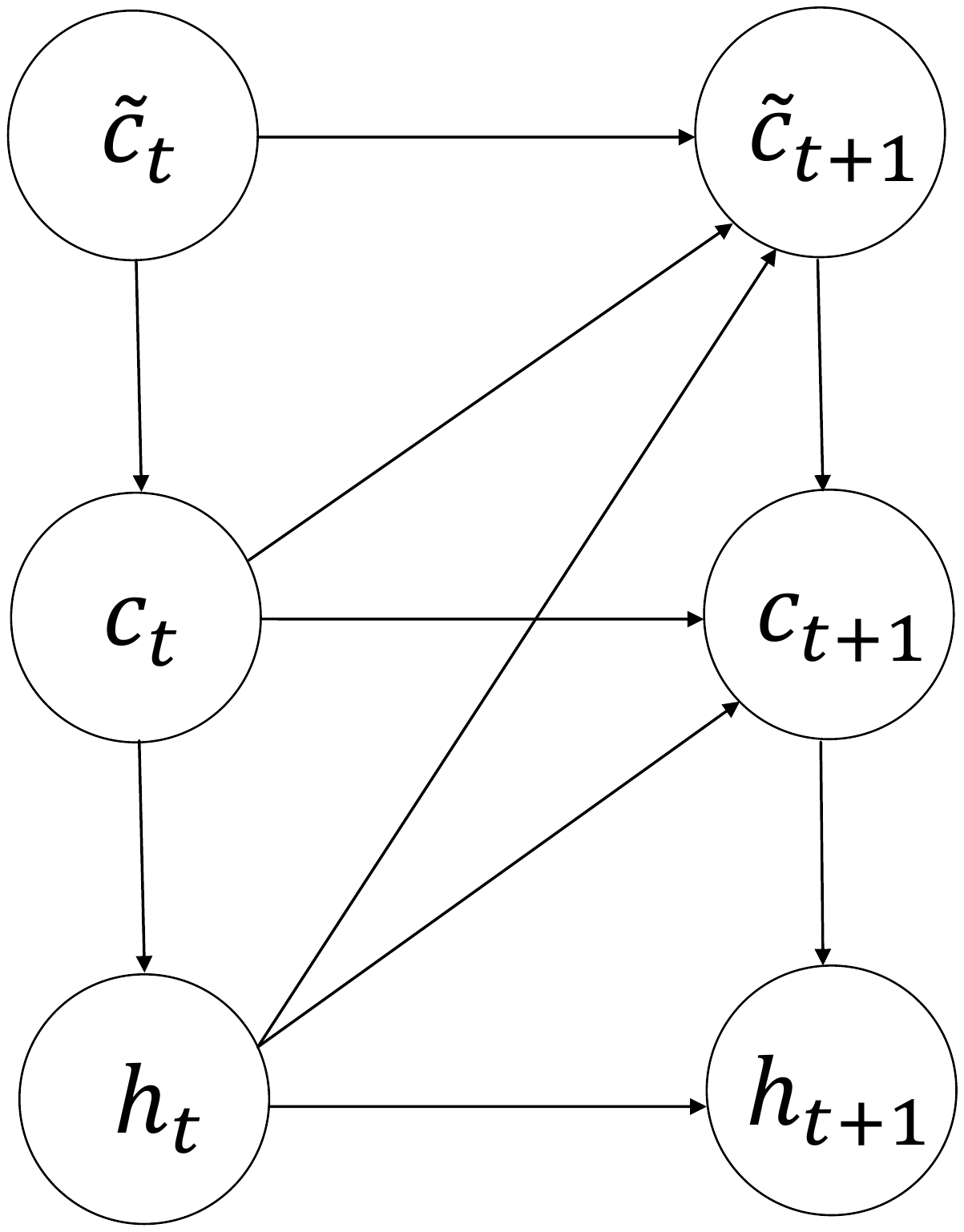}}
\caption{Computational graphs of the LSTM, Stacked LSTM and Nested LSTM.  The hidden state, outer memory cell, and inner memory cell are represented by $h$, $c$, and $d$, respectively.  While the current hidden state can influence the contents of the next inner memory cell directly, the inner memory influences the hidden state only via the outer memory.}
\label{fig:fig}
\end{figure*}

\subsection{The architecture}

In an LSTM, the equations updating the cell state and the gates are given by:
\begin{align}
        \boldsymbol{i}_t &= \boldsymbol{\sigma}_i(\boldsymbol{x}_t \boldsymbol{W}_{xi} + \boldsymbol{h}_{t-1} \boldsymbol{W}_{hi}
               + \boldsymbol{b}_i)\\
        \boldsymbol{f}_t &= \boldsymbol{\sigma}_f(\boldsymbol{x}_t \boldsymbol{W}_{xf} + \boldsymbol{h}_{t-1} \boldsymbol{W}_{hf}
               + \boldsymbol{b}_f)\\
        \boldsymbol{c}_t &= \boldsymbol{f}_t \odot \boldsymbol{c}_{t - 1} \\
               &+ \boldsymbol{i}_t \odot \boldsymbol{\sigma}_c(\boldsymbol{x}_t \boldsymbol{W}_{xc} + \boldsymbol{h}_{t-1} \boldsymbol{W}_{hc} + \boldsymbol{b}_c)\\
        \boldsymbol{o}_t &= \boldsymbol{\sigma}_o(\boldsymbol{x}_t \boldsymbol{W}_{xo} + \boldsymbol{h}_{t-1} \boldsymbol{W}_{ho} + \boldsymbol{b}_o)\\
        \boldsymbol{h}_t &= \boldsymbol{o}_t \odot \boldsymbol{\sigma}_h(\boldsymbol{c}_t)
\end{align}

Note that these equations are similar to those defined in \cite{gravesPTB}, but do not include peephole connections.

Nested LSTMs replace the addition operation used to compute $c_t$ in LSTMs with a learned, stateful function, $c_t = m_t(f_t \odot c_{t-1}, i_t \odot g_t
)$.
We refer to the state of the function, $m$ at time $t$ as the \emph{inner memory}, and calling the function to compute $c_t$ also computes $m_{t+1}$.
We chose to implement the memory function as another LSTM memory cell, producing a nested LSTM (see Figure~\ref{fig:nested} for an illustration).
The memory function could instead be another \emph{Nested} LSTM cell, permitting arbitrarily deep nesting.

Given these architecture choices, the input and the hidden states of the memory function in an NLSTM become:

\begin{align}
        \boldsymbol{\widetilde{h}}_{t-1} &= \boldsymbol{f}_t \odot \boldsymbol{c}_{t - 1} \\
        \boldsymbol{\widetilde{x}}_t &= \boldsymbol{i}_t \odot \boldsymbol{\sigma}_c(\boldsymbol{x}_t \boldsymbol{W}_{xc} + \boldsymbol{h}_{t-1} \boldsymbol{W}_{hc} + \boldsymbol{b}_c)
\intertext{In particular, note that if the memory function is addition, the entire system reduces to the classical LSTM, since the cell update becomes}
        \boldsymbol{c}_t &= \boldsymbol{\widetilde{h}}_{t-1} + \boldsymbol{\widetilde{x}}_t
\end{align}

In the architectural variant of the Nested LSTM proposed here, an LSTM is used as the memory function, and the working of the inner LSTM is governed by:

\begin{align}
        \boldsymbol{\widetilde{i}}_t &= \boldsymbol{\widetilde{\sigma}}_i(\boldsymbol{\widetilde{x}}_t \boldsymbol{\widetilde{W}}_{xi} +
                            \boldsymbol{\widetilde{h}}_{t-1} \boldsymbol{\widetilde{W}}_{hi} + \boldsymbol{\widetilde{b}}_i)\\
        \boldsymbol{\widetilde{f}}_t &= \boldsymbol{\widetilde{\sigma}}_f(\boldsymbol{\widetilde{x}}_t \boldsymbol{\widetilde{W}}_{xf} +
                            \boldsymbol{\widetilde{h}}_{t-1} \boldsymbol{\widetilde{W}}_{hf} + \boldsymbol{\widetilde{b}}_f)\\
\begin{split}
        \boldsymbol{\widetilde{c}}_t &= \boldsymbol{\widetilde{f}}_t \odot \boldsymbol{\widetilde{c}}_{t - 1} \\
        &+ \boldsymbol{\widetilde{i}}_t 
                            \odot \boldsymbol{\widetilde{\sigma}}_c(\boldsymbol{\widetilde{x}}_t \boldsymbol{\widetilde{W}}_{xc} + 
                            \boldsymbol{\widetilde{h}}_{t-1} \boldsymbol{\widetilde{W}}_{hc} + \boldsymbol{\widetilde{b}}_c)
\end{split}\\
        \boldsymbol{\widetilde{o}}_t &= \boldsymbol{\widetilde{\sigma}}_o(\boldsymbol{\widetilde{x}}_t \boldsymbol{\widetilde{W}}_{xo} +
                            \boldsymbol{\widetilde{h}}_{t-1} \boldsymbol{\widetilde{W}}_{ho} + \boldsymbol{\widetilde{b}}_o)\\
        \boldsymbol{\widetilde{h}}_t &= \boldsymbol{\widetilde{o}}_t \odot \boldsymbol{\widetilde{\sigma}}_h(\boldsymbol{\widetilde{c}}_t) \\
\intertext{The cell state update of the outer LSTM now becomes:}
        \boldsymbol{c}_t &= \boldsymbol{\widetilde{h}}_t        
\end{align}

\FloatBarrier
\section{Experiments}

We evaluate Nested LSTMs on a wide variety of datasets and tasks: the Penn Treebank Corpus \citep{PTB} and the larger Text8 dataset \citep{Text8} (both representing standard character-level language modeling, with Text8 being much larger than the Penn Treebank Corpus), the Chinese Poem Generation dataset \citep{zhang2014chinese} (which requires character-level language modeling on much smaller sequences with less temporal dependency than is common, but with a significantly larger number of characters than would typically be found in English), and the MNIST Glimpses task \citep{ba2016using} (which is a classification task, but one that contains temporal dependencies).  We show that, in spite of these tasks representing diverse scenarios and objectives, nested LSTMs consistently improve performance over corresponding stacked LSTM baselines with a comparable number of parameters.

We set ${\sigma}_i$, ${\sigma}_f$, ${\sigma}_o$, $\widetilde{{\sigma}}_i$, $\widetilde{{\sigma}}_f$ and $\widetilde{{\sigma}}_o$ to the $sigmoid$ activation function, $\widetilde{{\sigma}}_c$, $\widetilde{{\sigma}}_h$ and ${\sigma}_h$ to $tanh$, and ${\sigma}_c$ to the $identity$ function in all our Nested LSTM experiments.

In all the following experiments, we initialize the hyperparameters of the Nested LSTM and the stacked LSTM baselines identically. Although we explicitly specify the hyperparameters, unless otherwise mentioned, the hyperparameters we use are identical to those used in \citet{zoneout, rnn_batchnorm2}. We initialize the nested and stacked LSTMs' first input gates (which convert the input vector from having $vocabulary$ number of elements to $cell\_size$ number of elements) using a Glorot initialization scheme \cite{glorot2010understanding}, while all other gates in the LSTM (the other input, remember and output gates, the final output gate) are initialized using orthogonal initialization \cite{saxe2013exact}.

We also try to match the number of parameters of the different stacked LSTM baselines as closely as possible with our 2-layer Nested LSTM by adjusting the number of hidden units. While this is possible precisely with the 2-layer stacked baseline, this is slightly harder to achieve with a single-layered or 3-layered stacked LSTM. Correspondingly, we show the results of two single-layered LSTMs: one with the same number of parameters as in the reference paper, and one with a number of parameters \emph{larger} than those used by our model. We also choose the number of hidden units of the 3-layered stacked LSTM to surpass the number of parameters used by our model. Thus, our model has an equal number of parameters as the 2-layered stacked LSTM, and is at a \emph{disadvantage} to the larger single layered LSTM and the 3-layered stacked LSTM, but outperforms all these baselines. 

\subsection{Visualization} \label{sec:visualization}

To analyze what the cell activations look like and how they depend on each other, we visualize the changes in the cell activation states of both the inner and outer cell in the Nested LSTM as the sequence is fed into it. We do this on the model trained on the Penn Treebank dataset as described in Section~\ref{expt-ptb}, and then visualize the cell states as a sequence from the test set is fed into the Nested LSTM as in \citet{karpathy2015visualizing}.

\begin{figure*}[hbt]
	\centering
	\includegraphics[width=0.49\textwidth]{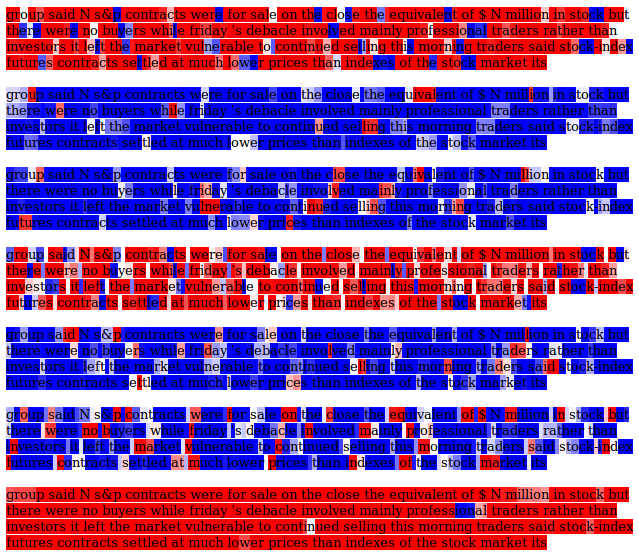}
	\hfill
	\includegraphics[width=0.49\textwidth]{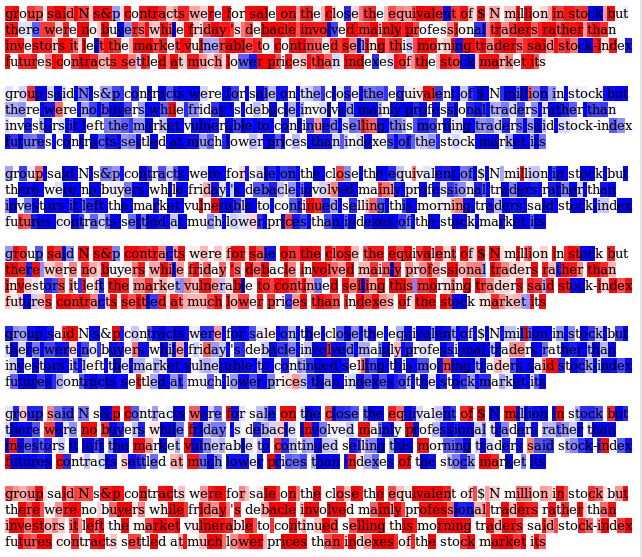}
	\caption{
		A visualization of the cell activations corresponding to the input character for the inner cell (left) and the outer cell (right). Red implies a negative cell state value, and blue a positive one. A darker shade implies a larger magnitude. In the case of the states of the inner LSTM, we visualize $tanh(\boldsymbol{\widetilde{c}}_t)$ (since $\boldsymbol{\widetilde{c}}_t$ is not constrained), while in the case of the outer LSTM, we directly visualize $\boldsymbol{c}_t$.}
	\label{fig:ptb-vis-cell}
\end{figure*}
\begin{figure*}[!t]
	\centering
	\includegraphics[width=0.49\textwidth]{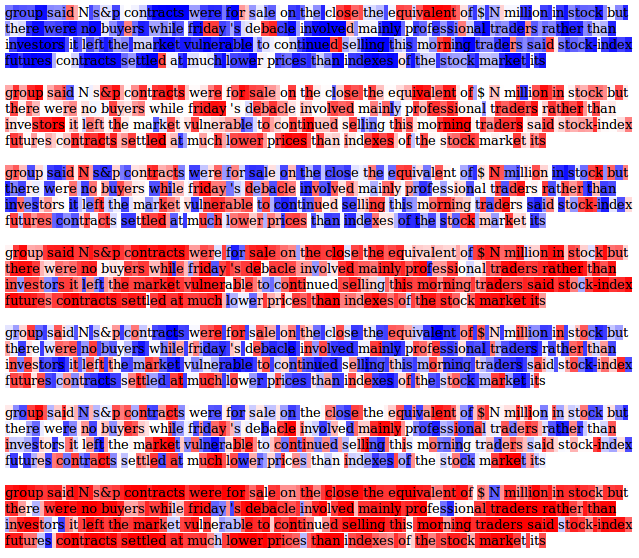}
	\hfill
	\includegraphics[width=0.49\textwidth]{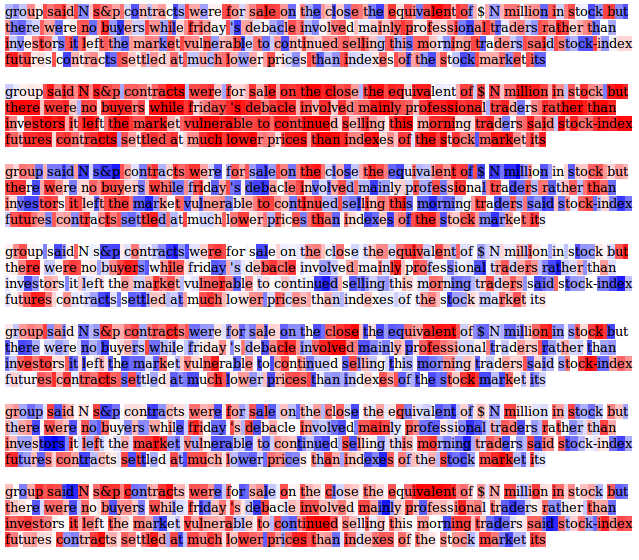}
	\caption{
		A visualization of $tanh(\boldsymbol{c}^n_t)$, representing the cell activations, corresponding to the input character for the first (right) and second (left) stacked layers. Red implies a negative cell state value, and blue a positive one. A darker shade implies a larger magnitude.}
	\label{fig:ptb-vis-stacked-cell}
\end{figure*}

We show our resulting visualization in Figure~\ref{fig:ptb-vis-cell}. 
The cells for which the visualizations have been shown are the first seven cells of the model. 
From the visualization, we see that the inner LSTM's cell activations tend to be relatively consistent across many time-steps, while the outer LSTM's cell activations fluctuate much more rapidly. 
This visualization demonstrates that the NLSTM hierarchy works as expected: outer memory operates at a shorter time-scale and uses inner memory to store longer-term information. 

We contrast this with a similar visualization of a 2-layer stacked LSTM baseline, in Figure~\ref{fig:ptb-vis-stacked-cell}. 
While the higher layer memory (which is "further away" from the input) operates at a longer time-scale than the lower layer memory, it still fluctuates more rapidly than the inner cells of the NLSTM.
This indicates that the NLSTM's ability to selectively process and remember information across multiple levels of nested memories frees the model to remember information over longer-periods, and supports our intuition that nested memory structures can form more effective temporal hierarchies.
\FloatBarrier

\subsection{Penn Treebank Character-level Language Modeling} \label{expt-ptb}

The Penn Treebank dataset \citep{PTB} contains around 1 million words, with a standard train:validation:test split. We train models on this dataset to perform character-level prediction, given an input sequence, and measure the negative log likelihood (NLL) loss and the bits per character (BPC, defined as the NLL divided by $ln 2$).

\begin{figure}[htb]
\centering
\includegraphics[width=0.45\textwidth]{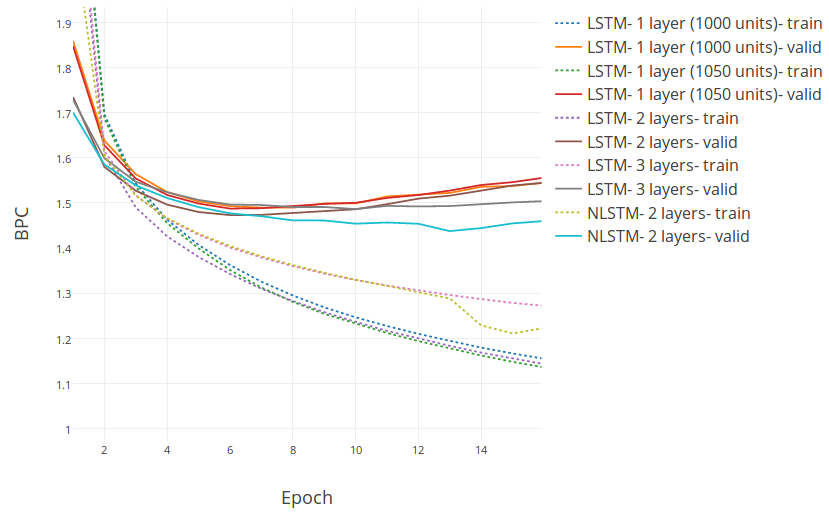}
\caption{
BPC as a function of epoch for character-level language modeling on PTB's test and validation sets}
\label{fig:charPTB}
\end{figure}

\begin{table}[h!]
\centering
\small
\begin{tabular}{@{}lllllll@{}}
\toprule
Model & n & cell size & Params & Valid & Test \\ \midrule
LSTM & 1 & 1000 & 4.25M & 1.489 & 1.451 \\
LSTM & 1 & 1050 & 4.68M & 1.487 & 1.448 \\
LSTM & 2 & 600 & 4.47M & 1.473 & 1.434 \\
LSTM & 3 & 450 & 4.17M & 1.486 & 1.448 \\
NLSTM & 2 & 600 & 4.47M & \textbf{1.437} & \textbf{1.399} \\
 \bottomrule
\end{tabular}
\caption{BPC Losses for the Nested LSTM versus various baselines. The test BPC losses correspond to the respective model's loss at the epoch in which it had the minimum valid BPC (also shown).}
\label{tab:ptb-results}
\end{table}

For Penn Treebank, our first baseline is a single layer LSTM of 1000 hidden units, following prior works \citep{gravesPTB,norm_stabilizer,zoneout, rnn_batchnorm2}.
We compare this architecture with 2-layer and 3-layer stacked LSTMs and 2-layer nested LSTMs. The number of hidden units of each model is chosen to (approximately) balance the capacity at around 4 million parameters. We also choose a single layered LSTM with a larger number of parameters than the 2-layered LSTM and NLSTM models.
We train using Adam \citep{adam} with a learning rate of 0.002 in sequences of 100 and batches of 32, and clip gradients with a threshold of 1, as in the aforementioned papers. However, we train on non-overlapping sequences, and without any normalization (which we believe could further improve these results). We train each model for 35 epochs.

We find that nested LSTMs yield an improvement of .035 BPC over stacked LSTMs using the same number of hidden units and layers, which, in turn, outperforms other baseline models. Notably, both models and the 3-layered stack LSTM outperform the single-layer network, suggesting that the common use of single-layer nets for this task is sub-optimal. Learning curves are presented in Figure~\ref{fig:charPTB}.

\FloatBarrier

\subsection{Chinese Poetry Generation}

\begin{table}[h!]
\centering
\small
\begin{tabular}{@{}lllllll@{}}
\toprule
Model & n & cell size & Params & Valid & Test \\ \midrule
LSTM & 1 & 32 & 868k & 680.15 & 669.99 \\
LSTM & 1 & 40 & 1.1M & 674.82 & 670.27 \\
LSTM & 2 & 32 & 877k & 810.54 & 771.88 \\
LSTM & 3 & 32 & 885k & 944.93 & 925.47 \\
NLSTM & 2 & 32 & 877k & \textbf{629.71} & \textbf{625.19} \\
 \bottomrule
\end{tabular}
\caption{Perplexity for the Nested LSTM versus various baselines on the Chinese Poetry Generation dataset}
\label{tab:charpg-results}
\end{table}

\begin{figure}[htb]
\centering
\includegraphics[width=0.45\textwidth]{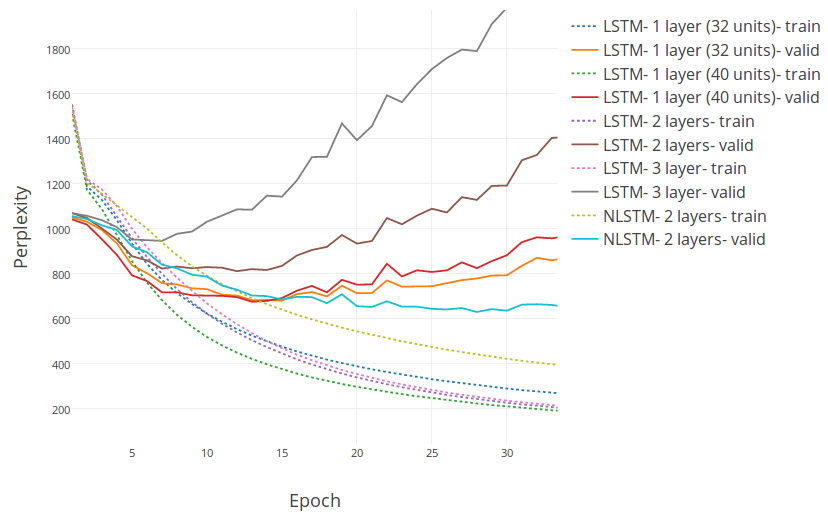}
\caption{
Perplexity as a function of epoch for character-level prediction on ChinesePG's train and validation sets}
\label{fig:charpg}
\end{figure}

Here, we use the subset of the Chinese Poem Generation dataset \cite{zhang2014chinese} comprised of quatrains with 5-characters each, with the standard specified train:validation:test split. This task is significantly different from the PTB task: the sequence length (and as a result the length of the temporal dependency) is much shorter, but the number of characters (over 5000) is \emph{two orders of magnitude} larger.

We follow the hyperparameter set used by the LSTM character-level prediction task used as the baseline in \citet{yu2017seqgan} (using a learning rate of 0.002). Following suit, we keep one of our baselines as single-layered LSTM with a cell size of 32. Our other baselines are a single layered LSTM with cell size 40, a 2-layered stacked LSTM with cell size 32 and a 3-layered stacked LSTM with cell size 32. We compare these baselines against our Nested LSTM with a cell size of 32. Note that because of the small cell sizes, the number of parameters in all these models is roughly the same (around 850k) in spite of the different numbers of layers (except in the case of the single layered LSTM with a cell size of 40, which has around 1.1M parameters). All models that we compare against, however, have more or equal parameters when compared to the Nested LSTM (except for the 32-cell single layered LSTM, which is why we introduce the additional 40-cell single layered baseline). We measure the performance using perplexity (which is $e^{NLL}$), as in \citet{che2017maximum}.

We find that the Nested LSTM outperforms all the baselines by a perplexity of \~45 on the test set. 
Surprisingly, we find that both the single-layered LSTM baselines outperform the corresponding stacked  LSTM baselines, even the one-layered LSTM with slightly fewer parameters. This is possibly because of the relatively small cell size used in this experiment. However, we observe that the  Nested LSTM outperforms the single-layered LSTM in this case as well, pointing to its robustness with respect to the model size (i.e., the number of cells, and by extension, parameters, used in the model).

\subsection{MNIST Glimpses}

\begin{table*}[!htb]
\centering
\small
\begin{tabular}{@{}lllllllll@{}}
\toprule
Model & n & cell size & Params & Valid NLL & Test NLL & Valid Accuracy (\%) & Test Accuracy (\%) \\ \midrule
LSTM & 1 & 100 & 61.0k & 0.1007 & 0.1229 & 97.85\% & 97.19\% \\
LSTM & 1 & 130 & 94.9k & 0.1070 & 0.1242 & 97.89\% & 97.51\% \\
LSTM & 2 & 75 & 83.6k & 0.1040 & 0.1149 & 98.15\% & 97.45\% \\
LSTM & 3 & 75 & 85.1k & 0.1077 & 0.1242 & 98.07\% & 97.46\% \\
NLSTM & 2 & 75 & 83.6k & \textbf{0.0836} & \textbf{0.1136} & \textbf{98.23\%} & \textbf{97.60\%} \\
 \bottomrule
\end{tabular}
\caption{NLL and percentage error for the Nested LSTM versus various baselines on the MNIST Glimpses task. The epoch at which the percentage error has been show corresponds to that at which each model had the lowest percentage error on the validation set. Similarly for NLL, the model's validation NLL has been used to determine the epoch at which the test NLL is examined.}
\label{tab:gmnist-results}
\end{table*}

\begin{figure}[htb]
\centering
\includegraphics[width=0.45\textwidth]{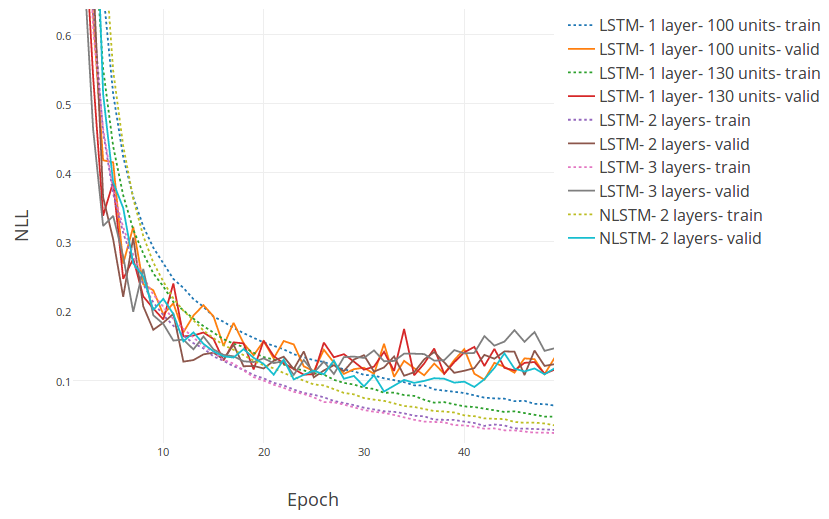}
\includegraphics[width=0.45\textwidth]{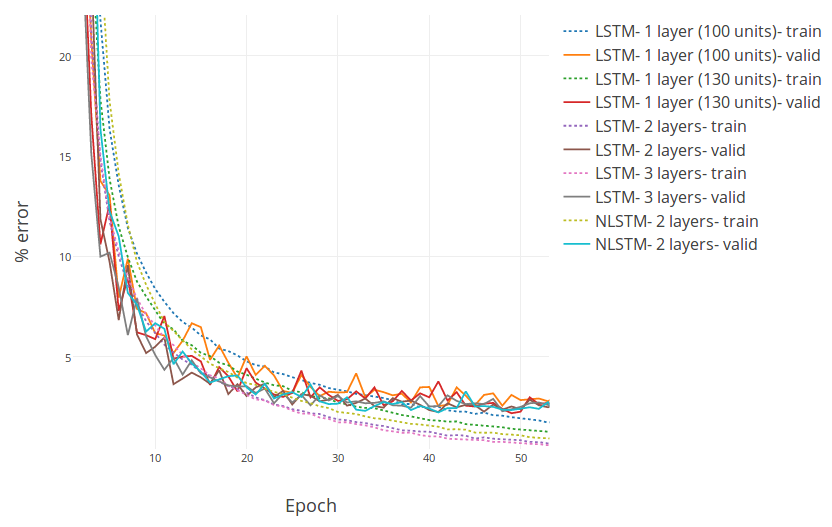}
\caption{
Plots of NLL (top) and percentage error (bottom) on the MNIST glimpses' train and validation sets (versus the epoch)}
\label{fig:gmnist}
\end{figure}

In the MNIST Glimpses task, introduced in \citet{ba2016using}, each 28x28 image (with the pixel values normalized to the range [0, 1]) is split into 4 quadrants. Glimpses of each quadrant (in the form of alternate rows and columns), followed by the entire quadrant are then fed sequentially into the model (with 20 elements in the sequence, each element comprising of 49 pixels). The model then predicts which integer the input represented.

The hyperparameter set we use is similar to that chosen in the \emph{pMNIST} task in \citet{zoneout}: we train all models with an RMS Prop optimizer \cite{rmsprop} with a learning rate of 0.001 for 150 epochs (note that here, we use the more commonly used decay rate of 0.9 instead of 0.5 used there), and clip the gradients to a maximum norm of 1. As in the aforementioned \emph{pMNIST} task, we use a 100 cell single layer LSTM baseline, along with 130 cell single layer, 75 cell two-layered stacked and 75 cell three-layered stacked LSTM baseline, and compare these baselines with a 75 cell Nested LSTM.

The Nested LSTM outperforms the (stacked) LSTM baselines in terms of both NLL and error percentage, both on the validation and test datasets. In particular, it reduces the validation error by 4.3\% when compared to the next best performing model (to 1.77\%, down from 1.85\% for a 2-layered stacked LSTM), and the validation NLL by almost \emph{17\%} (down to 0.0836 from 0.1007 in the case of a single-layered LSTM).

\FloatBarrier

\subsection{text8}

\begin{table}[htb]
	\centering
	\small
	\begin{tabular}{@{}lllllll@{}}
		\toprule
		Model & n & cell size & Params & Valid & Test \\ \midrule
		LSTM & 1 & 2000 & 16.28M & 1.399 & 1.482 \\
		LSTM & 1 & 2100 & 17.93M & 1.396 & 1.480 \\
		LSTM & 2 & 1200 & 17.45M & 1.385 & 1.466 \\
		LSTM & 3 & 950 & 18.19M & 1.389 & 1.471 \\
		NLSTM & 2 & 1200 & 17.45M & \textbf{1.363} & \textbf{1.445} \\
		\bottomrule
	\end{tabular}
	\caption{BPC for the Nested LSTM versus various baselines on the text8 task}
	\label{tab:text8-results}
\end{table}

\begin{figure}[htb]
\centering
\includegraphics[width=0.45\textwidth]{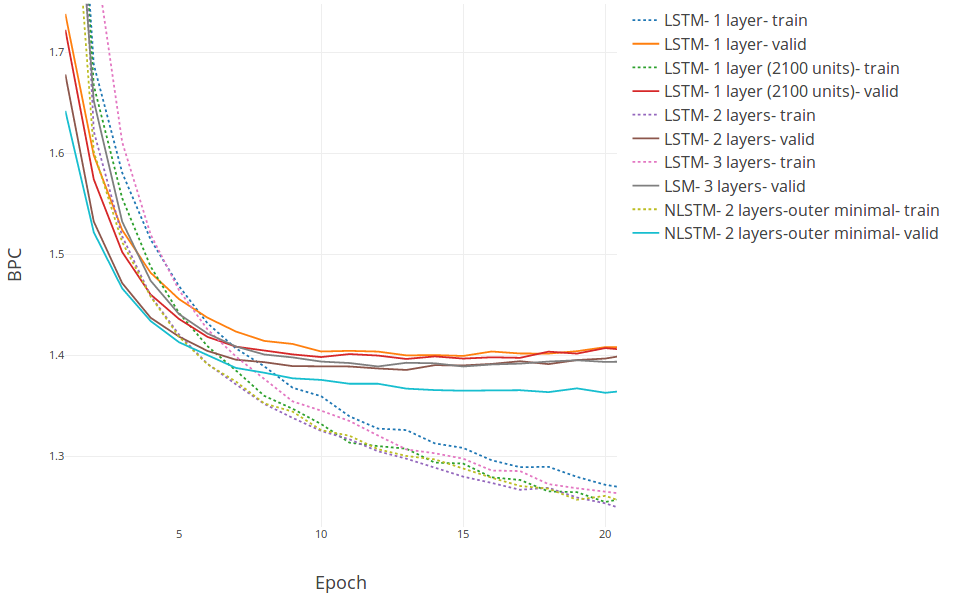}
\caption{
	BPC vs epoch curves for character-level prediction on text8's train and validation sets}
\label{fig:text8}
\end{figure}

The text8 dataset comprises of the first 100MB of a cleaned-up version of enwik8 (which is comprised of text from Wikipedia) \citep{Text8}.

As in our earlier experiments, we keep our hyperparameters identical to \citet{zoneout, rnn_batchnorm2}, except that we do not use any normalization, and that we train on non-overlapping sequences: we use a learning rate of 0.001, batch size of 128, sequence length of 180, a gradient clipping threshold of 1, with an adam optimizer \citep{adam}. Each model is trained for 40 epochs. Our baselines include 2000 and 2100 celled single layered LSTMs, a 1200 celled two-layered stacked LSTM and a 950 celled three-layered stacked LSTM baseline, pitted against a 1200 celled Nested LSTM.

Here too, we observe that our model outperforms the closest baseline (a 2-layered stacked LSTM) on both the valid (1.363 vs 1.385, respectively) and test (1.445 vs 1.466, respectively) sets. This indicates both that the proposed Nested LSTM is robust to different model sizes (as shown by the improvement it affords in the Chinese Poem generation task, where a relatively very small model was used), and that larger models trained on large datasets benefit from a nested architecture.

\FloatBarrier

\section{Conclusions}

%
Nested LSTMs (NLSTM) are a simple extension of the LSTM model that add depth via nesting, as opposed to via stacking.
The inner memory cells of an NLSTM form an internal memory, which is only accessible to other computational elements via the outer memory cells, implementing a form of temporal hierarchy.
NLSTMs outperform stacked LSTMs with similar numbers of parameters in our experiments, and result in more well defined temporal hierarchies in the activations of their memory cells compares with stacked LSTMs. Thus, NLSTMs represent a promising alternative to stacked models.

\FloatBarrier

\subsubsection*{Acknowledgments}
We thank Christopher Beckham for help with early experiments, and Tegan Maharaj and Nicolas Ballas for helpful discussions.
We thank the developers of Theano \citep{Theano}, Fuel, and Blocks \citep{Blocks}.
We acknowledge the computing resources provided by ComputeCanada and CalculQuebec.

\bibliography{reference}

\end{document}